\documentclass{ecai} 

\usepackage{latexsym}
\usepackage{amssymb}
\usepackage{amsmath}
\usepackage{amsthm}
\usepackage{booktabs}
\usepackage{enumitem}
\usepackage{graphicx}
\usepackage{color}

\newtheorem{definition}{Definition}

\newtheorem{assumption}{Assumption}

\usepackage{tikz}
\usetikzlibrary{bayesnet}
\usetikzlibrary{arrows}
\usetikzlibrary{backgrounds}

\makeatletter
\let\MYcaption\@makecaption
\makeatother

\usepackage{subfigure}

\makeatletter
\let\@makecaption\MYcaption
\makeatother

\DeclareMathOperator*{\varianceoperator}{\mathbb{V}}
\newcommand{\variance}[2]{\varianceoperator_{#1}\left[#2\right]}

\newcommand{\pt}{p^{(t)}(\bx,y)}

\newcommand{\bmu}{\bm{\mu}}
\newcommand{\bell}{\bm{\ell}}

\newcommand{\doubleB}{\mathbb{B}}
\newcommand{\expbnf}[2]{\exp\left( - \frac{#1}{#2}\right)}
\newcommand{\DKL}[2]{D_{\mathrm{KL}}\left( #1 \middle|\middle| #2 \right)}

\newcommand{\bgamma}{\bm{\gamma}}

\newcommand{\innerprod}[2]{\left\langle #1, #2 \right\rangle}

\newcommand{\Parameters}{\Statex \hspace{-3.78ex}\textbf{Parameters:}~}

\newcommand{\IndStatex}{\Statex\hspace{\algorithmicindent}}
\newcommand{\InddStatex}{\Statex\hspace{\algorithmicindent}\hspace{\algorithmicindent}}
\newcommand{\IndddStatex}{\Statex\hspace{\algorithmicindent}\hspace{\algorithmicindent}\hspace{\algorithmicindent}}

\newcommand{\algrule}[1][.2pt]{\par\vskip.3\baselineskip\hrule height #1\par\vskip.3\baselineskip}

\usepackage{algpseudocode}
\usepackage{algorithm, algpseudocode}

\newcommand\alglab[1]{\label{alg:#1}}
\renewcommand\algref[1]{Algorithm \ref{alg:#1}}

\algnewcommand\algorithmicforeach{\textbf{for each}}
\algdef{S}[FOR]{ForEach}[1]{\algorithmicforeach\ #1\ \algorithmicdo}
\algdef{SE}[DOWHILE]{Do}{DoWhile}[1]{\algorithmicdo\ #1}[1]{\algorithmicwhile\ #1}%

\algdef{SE}[DOUNTIL]{Do}{Until}[1]{\algorithmicdo\ #1}[1]{\algorithmicuntil\ #1}%

\usepackage{bm}

\let\originalleft\left
\let\originalright\right
\renewcommand{\left}{\mathopen{}\mathclose\bgroup\originalleft}
\renewcommand{\right}{\aftergroup\egroup\originalright}

\newcommand{\vone}{\nvector{1}}
\newcommand{\vzero}{\nvector{0}}
\newcommand{\mzero}{\nvector{0}}

\DeclareMathOperator{\diag}{diag}

\DeclareMathOperator{\softmax}{softmax}

\newcommand{\expb}[1]{\exp\left(#1\right)}
\newcommand{\logb}[1]{\log\left(#1\right)}

\DeclareMathOperator*{\expectoperator}{\mathbb{E}}
\newcommand{\expect}[2]{\expectoperator_{#1}\left[#2\right]}

\newcommand{\ndist}[2]{%
\mathcal{N}\left(#1, #2 \right)%
}

\newcommand{\func}[2]{ #1 \left(  #2 \right)}

\newcommand{\argmax}{\mathop{\rm arg~max}\limits}
\newcommand{\argmin}{\mathop{\rm arg~min}\limits}

\newcommand*{\norm}[1]{\left\| #1 \right\|}
\newcommand*{\enorm}[1]{\left\| #1 \right\|_2}
\newcommand*{\abs}[1]{\left| #1 \right|}
\newcommand*{\squared}[1]{\left( #1 \right)^2}

\newcommand*{\T}{\mathsf{T}}

\newcommand*{\nvector}[1]{\bm{\mathrm{#1}}}

\newcommand{\ba}{\nvector{a}}
\newcommand{\bb}{\nvector{b}}
\newcommand{\bc}{\nvector{c}}

\newcommand{\be}{\nvector{e}}

\newcommand{\bp}{\nvector{p}}

\newcommand{\bs}{\nvector{s}}

\newcommand{\bu}{\nvector{u}}

\newcommand{\bw}{\nvector{w}}
\newcommand{\bx}{\nvector{x}}

\newcommand{\bz}{\nvector{z}}

\newcommand{\mI}{\nvector{I}}

\newcommand{\Dtr}{D^{\mathrm{tr}}}

\newcommand{\Dte}{D^{\mathrm{te}}}

\newcommand{\Dva}{D^{\mathrm{va}}}

\usepackage{amssymb}
\usepackage{euscript}

\DeclareMathOperator{\doubleR}{\mathbb{R}}

\DeclareMathOperator{\doubleZ}{\mathbb{Z}}

\DeclareMathOperator{\scriptH}{\mathcal{H}}

\DeclareMathOperator{\scriptV}{\mathcal{V}}

\DeclareMathOperator{\scriptX}{\mathcal{X}}
\DeclareMathOperator{\scriptY}{\mathcal{Y}}

\newcommand\eqlab[1]{\label{eq:#1}}
\renewcommand*{\eqref}[1]{Eq. (\ref{eq:#1})}

\newcommand\seclab[1]{\label{sec:#1}}
\newcommand\secref[1]{Section \ref{sec:#1}}

\newcommand\figlab[1]{\label{fig:#1}}
\newcommand\figref[1]{{Figure \ref{fig:#1}}}
\newcommand\tabref[1]{{Table \ref{tab:#1}}}

\newcommand*{\deflab}[1]{\label{def:#1}}
\newcommand*{\defref}[1]{{Definition \ref{def:#1}}}

\newcommand\asslab[1]{\label{ass:#1}}
\newcommand\assref[1]{{Assumption \ref{ass:#1}}}

\let\supplementary\artsupplementary

\newcommand{\block}[1]{\smallskip\noindent{\textbf{#1}}}

\usepackage{enumitem} 
\newenvironment{slimitemize}{%
\begin{itemize}[label=\textbullet ,noitemsep,topsep=0pt,parsep=0pt,partopsep=0pt]}{\end{itemize}}

\usepackage{multirow}

\let\originalhline\hline
\renewcommand{\hline}{\originalhline \rule{0pt}{2.2ex}}

\newcommand{\bhm}[1]{\bm{\,\:#1^*}}
\newcommand{\chm}[1]{\bm{#1}}

\begin{document}


\begin{frontmatter}


\paperid{4625} 


\title{Source Component Shift Adaptation\\ via Offline Decomposition and Online Mixing Approach}


\author[A]{\fnms{Ryuta}~\snm{Matsuno}\orcid{0000-0002-4543-2128}\thanks{Email: ryuta-matsuno@nec.com}}
\address[A]{NEC Corporation}


\begin{abstract}
This paper addresses source component shift adaptation, aiming to update predictions adapting to source component shifts for incoming data streams based on past training data.
Existing online learning methods often fail to utilize recurring shifts effectively, while model-pool-based methods struggle to capture individual source components, leading to poor adaptation.
In this paper, we propose a source component shift adaptation method via an offline decomposition and online mixing approach.
We theoretically identify that the problem can be divided into two subproblems: offline source component decomposition and online mixing weight adaptation.
Based on this, our method first determines prediction models, each of which learns a source component solely based on past training data offline through the EM algorithm.
Then, it updates the mixing weight of the prediction models for precise prediction through online convex optimization.
Thanks to our theoretical derivation, our method fully leverages the characteristics of the shifts, achieving superior adaptation performance over existing methods.
Experiments conducted on various real-world regression datasets demonstrate that our method outperforms baselines, reducing the cumulative test loss by up to 67.4\%.
\end{abstract}

\end{frontmatter}




\section{Introduction}
\begin{figure}[t]
\centering
\subfigure[Data at $t$ (SC1:SC2:SC3 = 6:1:3)]{\includegraphics[width=0.49\linewidth]{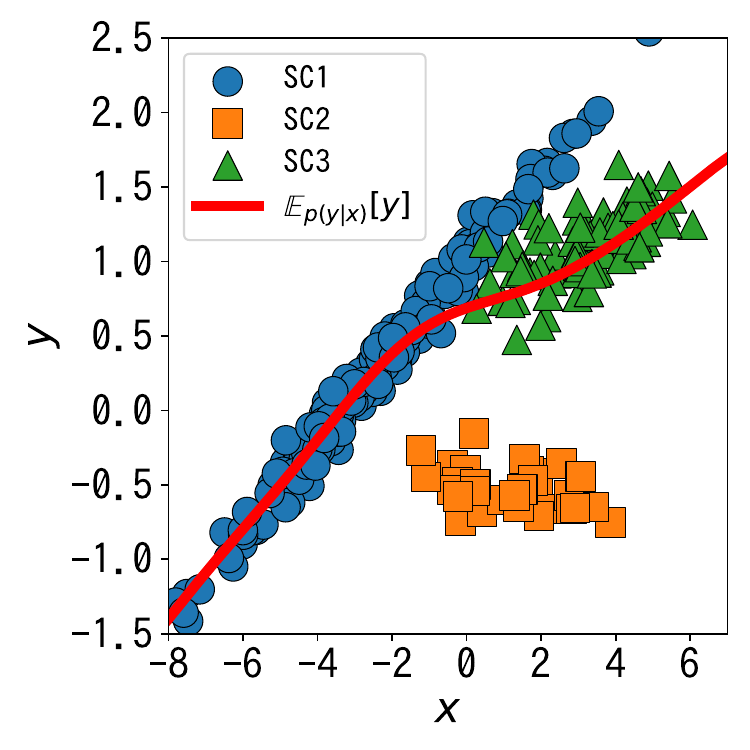}}
\subfigure[Data at $t'$ (SC1:SC2:SC3 = 1:6:3)]{\includegraphics[width=0.49\linewidth]{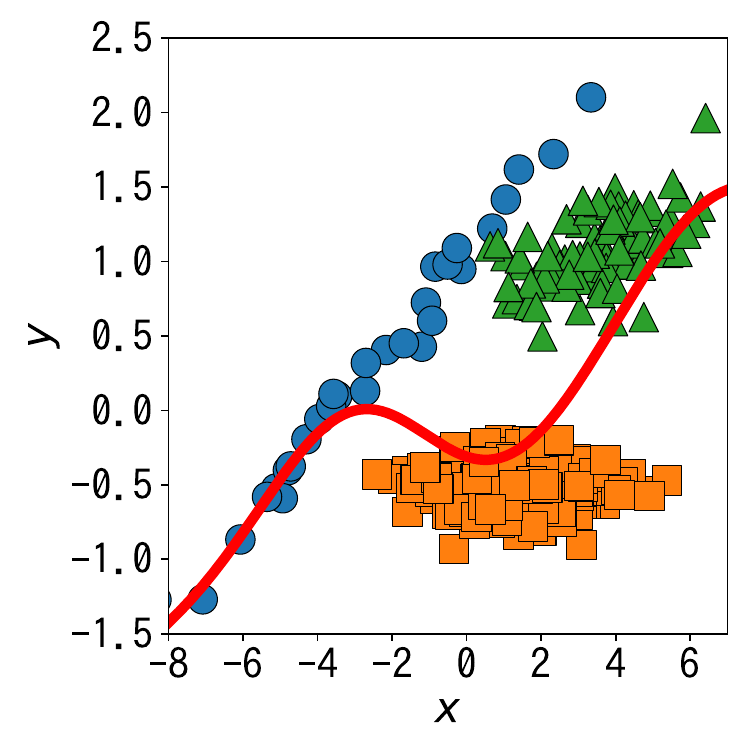}}
\caption{Illustration of source component shift with three Source Components (SC1 to SC3).
Red lines indicate $\expect{p(y|x)}{y}$, i.e., the best regression models.
As mixing weights of SC1 and SC2 change, the best models vary over time.}
\label{fig:scs}
\end{figure}

Data drift (a.k.a. data shift) adaptation  aims to maintain accurate predictions for drifting data streams using past training data and has demonstrated significant success across various industries.
A key type of data shift is the \textit{source component shift}~\cite{datasetshift2009}, characterized by gradual recurring drifts where multiple source components remain fixed, while their mixing weights change over time (\figref{scs}).
Examples of source component shifts include products supplied by different factories with varying proportions or sales demand shifts due to changes in the contribution of purchasers.

While \textit{``source component shift may be the most common form of dataset shift''}~\cite{datasetshift2009}, there are few researches studying source component shifts, limiting the applicability of recent advanced machine learning (ML) methods to this type of shifts.
Hence, this paper investigates source component shift adaptation, aiming to update predictions for incoming data streams considering source component shifts, given past training data.
In addition, we focus on regression tasks since regression highlights changes in data distributions more directly than classification and requires precise adaptation methods.

Naively, we can apply online learning methods, represented by online gradient descent (OGD) to source component shift adaptation.
However, these methods fail to exploit the recurring nature of shifts, limiting their effectiveness.
Model-pool methods~\cite{sea2001,kolter2007dwm,wang03awe,aue2,zhao18condor,song22sega,deckert11bwe,chu04aboost,elwell11learnppnse,graphpool2018} attempt to handle shifts by training new models and adding them to a pool for improved adaptation.
However, they lack explicit data decomposition aligned with source components, which is essential for accurate adaptation to source component shifts~\cite{datasetshift2009}.
Although FUZZ-CARE~\cite{fuzz2020song} incorporates data decomposition, its heuristic-based modeling, using naive Euclidean distances, lacks robust theoretical support, limiting adaptation to source component shifts.

In this paper, we propose a source component shift adaptation method via an offline decomposition and online mixing approach.
We first theoretically divide the problem into two distinct subproblems: offline source component decomposition and online mixing weight adaptation.
On the basis of this, our method first decomposes and learns prediction models corresponding to each source component based solely on past training data via the EM algorithm~\cite{em1977}.
It then adapts the mixing weight (i.e., ensemble weights) of these models through online convex optimization.
Since our method is based on  theoretical derivation that fully exploits the nature of source component shifts, it accurately adapts to the shifts, enhancing its prediction performance.
In addition,  since our proposed method prepares prediction models offline, it is rather robust against online noises and requires less running costs compared to the conventional model pool-based methods.
We perform numerical experiments on diverse real-world-based regression datasets to verify the effectiveness of our method.
Results show that our method achieves superior prediction performance over the baselines including FUZZ-CARE, reducing cumulative test loss by up to 67.4\%.
Finally, our contributions are summarized as follows:
\begin{slimitemize}
\item We theoretically divide the source component shift adaptation problem into offline source component decomposition and online mixing weight adaptation (Section 3.1).
\item We propose a source component shift adaptation method 
based on the theoretical division of the problem with the EM algorithm and online convex optimization (Section 3.2).
\item We conduct empirical evaluation and show that our method demonstrates superior prediction performance, significantly reducing cumulative test loss over baselines (Section 4).
\end{slimitemize}

\section{Preliminary}

In this section, we briefly introduce the problem formulation of the source component shift adaptation and related works.

\subsection{Problem Formulation}
We consider a supervised regression task where the input space is $\scriptX \subseteq \doubleR^d$, $\doubleR$ represents the space of real values, $d \in \doubleZ_{\geq 1}$ is a positive integer indicating the dimensionality of the input space, and $\doubleZ_{\geq 1}$ is the space of integers no less than 1.
The output space is $\scriptY \subseteq \doubleR$.
In conventional regression tasks, a training dataset $D := \{(\bx_i, y_i)\}_{i=1}^N$ is given, consisting of $N$ i.i.d. samples drawn from a density function $p(\bx, y)$ over $\scriptX \times \scriptY$.
The objective is to train a prediction model $h: \scriptX \to \scriptY$ to minimize the expected squared error $R(h)$, defined as
\begin{align}
R(h) := \expect{p(\bx,y)}{\squared{h(\bx) - y}} . \eqlab{ese}
\end{align}

In this paper, we address a key type of data shifts known as \textit{source component shift}~\citep{datasetshift2009}, where multiple underlying source components exist  behind data generation and their mixing weight change over time.
Formally, source component shift is defined as follows.
\begin{definition}[Source component shift]
\deflab{scs} Let $\{p_k\}_{k=1}^K$ represent $K \in \doubleZ_{\geq 2}$ distinct densities over $\scriptX \times \scriptY$, called \textit{source components}.
Source component shift is a change of joint distribution over time $t$ where the density at time $t$ is formulated as
\begin{align}
p^{(t)}(\bx, y) &:= \sum_{k=1}^K w^{(t)}_k p_k(\bx, y),     \eqlab{scs}
\end{align}
where $\bw^{(t)} = \big[w_1^{(t)},\ldots,w_K^{(t)}\big]^\T \in \Delta^{K-1}$ is the mixing weight of the source components and $\Delta^{K-1} := \{ \bx \in [0,1]^K \mid \norm{\bx}_1 = 1 \}$ is the $K$-dimensional probability simplex.
\end{definition}
Similar to the above conventional regression, we are given past $N$ training samples, denoted as $D = \{(\bx_t, y_t)\}_{t=1}^N $, where $(\bx_t, y_t) \sim p^{(t)}(\bx, y)$.
Our goal is to predict outputs given inputs from $t=N+1$ until $t=N+T$.
Let $f_t:\scriptX \to \scriptY$ be a prediction model for time $t$.
The evaluation metric of $\{f_t\}_{t=N+1}^{N+T}$ is defined as a natural extension of \eqref{ese} as
\begin{align}
R(\{f_t\}) &:= \sum_{t=N+1}^{N+T} \expect{p^{(t)}(\bx, y)}{ \squared{f_t(\bx) - y} },     \eqlab{scsa}
\end{align}
and in practice, we use its empirical version:
\begin{align}
\widehat{R}(\{f_t\}; \Dte) &:= \sum_{t=N+1}^{N+T} \squared{f_t(\bx_t) - y_t},     \eqlab{scsap}
\end{align}
where $\Dte = \{(\bx_t, y_t)\}_{t=N+1}^{N+T}, (\bx_t,y_t) \sim \pt$ is the test samples.
Please note that we omit subscripts (e.g., $t=N+1$) and superscripts (e.g., $N+T$) of $\{f_t\}_{t=N+1}^{N+T}$ in equations for better readability.
Furthermore, regarding the training data $D$, we essentially assume that the mixing weight of each source component is not identically zero among $t \in \{1,...,N\}$, formulated as \assref{nonzero}.
\begin{assumption}
\asslab{nonzero}
$\forall k \in [K],  \exists t \in [N], w_k^{(t)} > 0$.
\end{assumption}
\noindent
We use $[K]$ for the set of integers from $1$ to $K$, i.e., $[K]:=\{1,\ldots,K\}$.
Note that if \assref{nonzero} does not hold, the data shift may not be considered as a source component shift.

For every $t \in \{N+1,\ldots,N+T\}$, we aim to find the optimal $f_t$ that minimizes $\expect{p^{(t)}(\bx, y)}{ \squared{f_t(\bx) - y} }$, using the training data $D$ and the observed data $\{(\bx_s,y_s)\}_{s=N+1}^{t-1}$ in order to minimize the overall loss $R(\{f_t\})$.
More formally, the problem we study is formulated as follows.
\begin{definition}[Source component shift adaptation]
\deflab{scsa}
Under \defref{scs} without knowing $K$, assuming \assref{nonzero}, and given a training data $D := \{ (\bx_t, y_t) \}_{t=1}^N$, where each $(\bx_t, y_t)$ is independently sampled from  $p^{(t)}(\bx, y)$, 
the problem is to minimize \eqref{scsa} (or practically \eqref{scsap}) by optimizing $f_t$ for every $t \in \{N+1,\ldots,N+T\}$  using available information prior to each prediction.
\end{definition}


\subsection{Related Works}
In what follows, we review model-pool based drift adaptation methods, which are more promising and suited for handling recurring drifts (including source component shifts) rather than other online learning method~\cite{reccurentsurvey2024}.

Model-pool based drift adaptation methods maintain multiple models in their model pools to improve prediction accuracy~\cite{ensemble2017survey}.
These methods have been studied for more than two decades~\cite{sea2001}, and are categorized into passive and active approaches~\cite{ensemble2017survey}.
Passive methods train a new model and add it to their model-pool for every observed batch/chunk (i.e., a specific number of samples)~\cite{kolter2007dwm,sea2001,wang03awe,elwell11learnppnse,aue2,song22sega,fuzz2020song}.
Active methods, however, rely on drift detection to optimize when to train new models, which depends on accurate detection~\cite{zhao18condor,chu04aboost,deckert11bwe}.
Different strategies are employed for training models.
While basic methods apply standard supervised learning~\cite{wang03awe,song22sega,deckert11bwe,aue2}, others adjust their training based on model accuracy within the pool~\cite{elwell11learnppnse,chu04aboost} or reuse model parameters~\cite{zhao18condor}.
The calculation of ensemble weights also varies, including uniform weights~\cite{sea2001,song22sega}, reducing weights for inaccurate models~\cite{kolter2007dwm,zhao18condor,kolter2005addexp,aue2}, and prioritizing recent models~\cite{wae2013}.
For further adaptation to recurring drifts, several techniques have been studied,  leveraging the similarities of training samples~\cite{rcd2013,redlla2012}, prediction performance~\cite{cpf2016}, and other meta information~\cite{meta2018,graphpool2018,wu2022}.
Survey studies offer further insights into further details~\cite{ensemble2017survey,reccurentsurvey2024}.

Most of these methods deal with classification tasks, relying on the discreteness of outputs~\cite{wang03awe,aue2,cpf2016}, and are not suited for regression tasks.
In addition, while source component shift is characterized by a mixture of multiple source components (i.e., concepts), and it is essential to learn each of the source components/concepts for optimal adaptation, most existing methods do not explicitly decompose data relative to the source components.
This makes these methods incapable to maintain effective model pools.

FUZZ-CARE~\cite{fuzz2020song} may be the most suitable method for source component shift adaptation and is the closest work to our proposing method.
It learns multiple centroids of inputs and the Euclidean distances to each centroid affect the learning of ensemble weights of prediction models.
The prediction models are trained to minimize prediction errors within the ensemble.
Its centroid-based clustering can be regarded as an explicit data decomposition, leading better learning of the model-pool for adapting to source component shifts.
FUZZ-CARE also supports online updates to models and ensemble weights for enhanced adaptability.
However, as the derivation of FUZZ-CARE is heuristic, its theoretical foundation remains uncertain.
Moreover, its naive use of Euclidean distances may not effectively decompose data corresponding to the true source components, especially for high-dimensional nonlinear data.

In~\cite{datasetshift2009}, a single-time source component shift adaptation is introduced based on the EM algorithm~\cite{em1977}.
However, this method is limited to only two source components and cannot be applied to general streaming data.

\section{Proposed Method}
\seclab{method}

In this section, we propose our method for source component shift adaptation.
We first reveal that source component shift adaptation can be divided into two distinct subproblems: offline source component decomposition and online mixing weight adaptation in \secref{prodeco}.
We then solve each subproblem in \secref{scd} and \secref{MWA}.



\subsection{Problem Decomposition}
\seclab{prodeco}

The expected form of the objective defined in \eqref{scsa} is rewritten as
\begin{align}
    R(\{f_t\}) 
    &= \sum_{t=N+1}^{N+T} \expect{p^{(t)}(\bx)}{\squared{f_t(\bx) - \expect{p^{(t)}(y|\bx)}{y}}} + C, \eqlab{u1}
\end{align}
where $C := \sum_{t=N+1}^{N+T} \expect{p^{(t)}(\bx)}{\variance{p^{(t)}(y|\bx)}{y} } $ is independent from $\{f_t\}_{t=N+1}^{N+T}$ and $\variance{p^{(t)}(y|\bx)}{y}  := \expect{p^{(t)}(y|\bx)}{\squared{y - \expect{p^{(t)}(y|\bx)}{y}}}$ is the variance of $y$ over $p^{(t)}(y|\bx)$.
By \defref{scs}, the conditional distribution $p^{(t)}(y|\bx)$ is derived as
\begin{align}
    &p^{(t)}(y|\bx)
    = \frac{p^{(t)}(\bx, y)}{p^{(t)}(\bx)} 
    =\sum_{k=1}^K \frac{w^{(t)}_k p_k(\bx)}{\sum_{k'=1}^K w^{(t)}_{k'} p_{k'}(\bx)} p_k(y|\bx) \\
    &~= \sum_{k=1}^K r_k^{(t)}(\bx) p_k(y|\bx) = r^{(t)}(\bx)^\T \left[p_1(y|\bx),...,p_K(y|\bx)\right]^\T \eqlab{cond}
\end{align}
where we define the mixing ratio for the $k$-th conditional distributions $r^{(t)}_k: \scriptX \to [0,1]$ as 
\begin{align}
r_k^{(t)}(\bx)  &:= \frac{w^{(t)}_k p_k(\bx)}{\sum_{k'=1}^K w^{(t)}_{k'} p_{k'}(\bx)} 
\end{align}
and $r^{(t)}: \scriptX \to \Delta^{K-1}$ can be written as
\begin{align}
r^{(t)}(\bx)  &:= \func{\softmax}{\log \bw^{(t)} + \log [p_1(\bx),...,p_K(\bx)]^\T}, \eqlab{defgt} 
\end{align}
where $\softmax: \doubleR^K \to \Delta^{K-1}$ computes the softmax among its input, i.e., $\left(\softmax(\bx)\right)_k := {\expb{x_k}}/{\sum_{k'} \expb{x_{k'}}}$.

Combining \eqref{u1} with \eqref{cond}, we have
\begin{align}
    R(\{f_t\}) 
    &= \sum_{t=N+1}^{N+T} \expect{p^{(t)}(\bx)}{\squared{f_t(\bx) - r^{(t)}(\bx)^\T \bm{\mu}(\bx) }} + C, \eqlab{u2}
\end{align}
where $\bmu(\bx) := [\mu_1(\bx),...,\mu_K(\bx)]^\T \in \scriptY^K$ is the vector of $\mu_k(\bx)$ and $\mu_k(\bx) := \expect{p_k(y|\bx)}{y}$ is the mean of the conditional distribution of $k$-th source component.
\eqref{u2} indicates that the minimization of $R(\{f_t\})$ corresponds to fit $f_t$ to $r^{(t)}(\bx)^\T \bm{\mu}(\bx)$.
This motivates us to model $f_t : \scriptX \to \scriptY$ by
\begin{align}
    f_t(\bx) :=  \func{\softmax}{\bu_t + v(\bx)}^\T h(\bx), \eqlab{fdef}
\end{align}
where $h: \scriptX \to \scriptY^{K}$ learns $\bmu$ and 
the modeling of $\func{\softmax}{\bu_t + v(\bx)}$ is motivated by \eqref{defgt}; this term learns the mixing ratio $r^{(t)}(\bx)$ with the weight $\bu_t \in \doubleR^K$ and the mapping function $v:\scriptX \to \doubleR^K$, indicating which source component would be active for an input $\bx_t$.
In particular, we approximate each marginal distribution of the source components (i.e., $p_k(\bx)$) by a normal distribution, resulting in modeling the mapping function $v$ as log of a density of a normal distribution.
Specifically, we model $v(\bx)_k$ as 
\begin{align}
    v(\bx)_k &:= \logb{ \frac{1}{\sqrt{(2\pi)^d \abs{\Sigma_k} }} \expb{ - \frac{(\bx - \bc_k)^\T \Sigma_k^{-1} (\bx - \bc_k)  }{2}}} \\
    &= - \frac{d \logb{2\pi}}{2} - \sum_{i=1}^d \log s_{ki} - \frac{\left((\bx - \bc_k)^2\right)^\T \bs_k^{-2} }{2}, \eqlab{normalv}
\end{align}
where $\bc_k \in \doubleR^d$ is the learnable mean of the normal distribution and $\Sigma_k := \diag(\bs_k^2) \in \doubleR_{>0}^{d \times d}$ with its learnable parameter $\bs_k = [s_{k1},...,s_{kd}] \in \doubleR^d_{> 0}$ is the covariance matrix.
However, it should be note that the modeling of $v$ can be any beside \eqref{normalv} and proper modeling may improve the overall performance.

This modeling of $f_t$ in \eqref{fdef} clearly separate the input-dependent part, consisting $h$ and $v$, from the time-dependent part $\bu_t$.
This means that although the time-dependent part should be maintained in the online adaption phase from $t = N+1$ to $t = N +T$, the non-time-dependent (i.e., input-dependent) part can be fitted and fixed offline using the past training data $D$ thanks to \assref{nonzero}.
On the basis of this observation, we decompose the problem of source component shift adaptation into two distinct subproblems as follows.
\begin{definition}[Source component decomposition]
\deflab{scd}
Given a training data $D := \{ (\bx_t, y_t) \}_{t=1}^N$, and two hypothesis spaces $\scriptH$ and $\scriptV$ for prediction models and mapping functions, respectively, the problem is to find $h^* \in \scriptH$ and $v^* \in \scriptV$ defined as 
\begin{align}
    (h^*, v^*) := \argmin_{(h, v \in \scriptH \times \scriptV)} \min_{\{\bu_t\}_{t=1}^N} R_N(\{f_t\}),
\end{align}
where $f_t$ is modeled with $h,v$, and $\bu_t$ by \eqref{fdef} and $R_N(\{f_t\})$ is the expected squared error among $t \in \{1,...,N\}$, defined as
\begin{align}
    R_N(\{f_t\}) := \sum_{t=1}^{N} \expect{p^{(t)}(\bx, y)}{ \squared{f_t(\bx) - y} }.
\end{align}
\end{definition}
\begin{definition}[Mixing weight adaptation]
\deflab{mwa}
Given a prediction model $h: \scriptX \to \scriptY^K$ and a mapping function $v:\scriptX \to \doubleR^K$, the problem is to minimize the following cumulative squared loss over test data $\Dte =\{(\bx_t, y_t)\}_{t=N+1}^{N+T}$, defined as $L(\{\bu_t\})$ as
\begin{align}
    L(\{\bu_t\})
    &:= \widehat{R}(\{f_t\}; \Dte)
    = \sum_{t=N+1}^{N+T} \squared{f_t(\bx_t) - y_t} 
    \eqlab{hatR}
\end{align}
by updating the weight $\bu_t \in \doubleR^K$ in \eqref{fdef} using $\{(\bx_s, y_s)\}_{s=N+1}^{t-1} $ for every $t=N+1,...,N+T$.
\end{definition}
\noindent
It should be remarked that we employ $\widehat{R}(\{f_t\}; \Dte)$ instead of $R(\{f_t\})$ in order to align with the evaluation used in related online learning studies~\cite{oogd2023}.
We solve each of subproblems in the subsequent \secref{scd} and \secref{MWA}, respectively.

\begin{figure}[t]
\begin{center}
\tikz{
\node[obs                           ] (y) {$y_t$};%
\node[latent,left=of y, yshift=-1cm ] (z) {$\bz_t$};%
\node[obs, above =of z           ] (x) {$\bx_t$};%
\node[const,right=of y,inner sep=.1cm] (h) {$h$};%
\node[const,left=of x,inner sep=.1cm] (u) {$\bu_t$};%
\node[const,left =of z,xshift=-1.2cm, inner sep=.1cm] (v) {$v$};%

\plate [inner sep=.3cm] {plate2} {(x)(y)(z)(u)} {$t\in[N]$}; %
\edge {x} {y}
\edge {z} {y}
\edge {x} {z}
\edge {u} {z}
\edge {v} {z}
\edge {h} {y}
}
\end{center}
\caption{Graphical model at prediction phase. The input $\bx_t \in \scriptX$ and its output $y_t \in \scriptY$ are observable, while the variable $\bz_t \in \doubleB^K$, which indicates which source component generated $\bx_t$ and $y_t$, is latent. Two functions $h:\scriptX \to \scriptY^K$ and $v:\scriptX \to \doubleR^K$ and weights $\{\bu_t\}_{t=1}^N \in \doubleR^{K\times N}$ are learnable.  }
\figlab{graphicalmodel}
\end{figure}

\subsection{Source Component Decomposition}
\seclab{scd}
In this section, we focus on solving \defref{scd} using maximum likelihood estimation (MLE).
By considering the likelihood, we are able to examine the distribution of the data $D$.
Existing FUZZ-CARE~\cite{fuzz2020song}, on the other hand, relies on squared errors of prediction, which are insufficient to characterize the data distribution.
Hence, MLE enables more accurate decomposition of each source component compared to FUZZ-CARE.

During the prediction phase, given a sample $(\bx_t, y_t) \in \scriptX \times \scriptY$, we do not know which source component generated it.
Let $k \in [K]$ be that source component, where $K \in \doubleZ_{\geq 2}$ is arbitrarily fixed for now and will be tuned in \secref{Ktune}.
The corresponding encoding of $k$ is denoted as $\bz_t = \be_k \in \doubleB^K$, where $\be_k \in \{0,1\}^{K}$ is a unit vector with all elements as 0 except the $k$-th element, which is 1, and the set $\doubleB^K = \{\be_1, \ldots, \be_K \}$ represents the standard basis of $\doubleR^K$.
Then, the latent vector $\bz_t$ is influenced by the mixing ratio $r^{(t)}(\bx_t) \in \Delta^{K-1}$, which is calculated with $\bu_t \in \doubleR^{K}$ and $v(\bx_t) \in \doubleR^K$.
The output $y_t$ is determined by $\bx_t$ and $\bz_t$ using the prediction model $h:\scriptX \to \scriptY^K$.
These relationships are illustrated as a graphical model in \figref{graphicalmodel}.

Our aim is to maximize the posterior $p(y_t| \bx_t, h, v, \bu_t)$, i.e., the likelihood of $h,v$, and $\bu_t$, computed as 
\begin{align}
    &p(y_t| \bx_t, h, v, \bu_t)
    = \sum_{\bz_t \in \doubleB^K} p(y_t| \bz_t, \bx_t, h) p(\bz_t | \bx_t, v, \bu_t) \\
    &= \frac{1}{\sqrt{2\pi \sigma^2}}\expbnf{\squared{y_t - h(\bx_t)}}{2\sigma^2}^\T \func{\softmax}{\bu_t + v(\bx_t)},
\end{align}
where $\sigma \in \doubleR_{> 0}$ is the noise level of outputs.
A standard solution for MLE with a latent variable (i.e., $\bz_t$) is the EM algorithm~\citep{em1977}.
It utilizes the following equality for arbitrary distribution $q$ over $\doubleB^K$;
\begin{align}
    \log p(y_t|\bx_t, h, v, \bu_t) &= F(q) + \DKL{q(\bz_t)}{ p(\bz_t|\bx_t, y_t, h,v,\bu_t)}, \eqlab{em}
\end{align}
where
\begin{align}
    \quad F(q) &:= \sum_{\bz_t \in \doubleB^K} q(\bz_t) \log \frac{p(y_t,\bz_t \mid \bx_t, h,v, \bu_t)}{q(\bz_t)},
\end{align}
and $\DKL{p}{q} := \sum_{\bz \in \doubleB^K} p(\bz) \log {\frac{p(\bz)}{q(\bz)}} $ computes the Kullback-Leibler divergence.
The EM algorithm repeat the subsequent E- and M-steps iteratively until convergence.
The overview of the obtaining algorithm is summarized in \algref{scd}.

\begin{algorithm}[t]
\caption{Source Component Decomposition. }
\alglab{scd}
\begin{algorithmic}[1]
\Require Training data $D = \{(\bx_t, y_t)\}_{t=1}^N$,
\IndStatex Number of components $K \in \doubleZ_{\geq 2}$,
\IndStatex Noise level $\sigma>0$,
\Parameters Learning rate $\eta > 0$ (default $0.01$),
\IndStatex Smoothing rate $\alpha > 0$ (default $0.1$),
\IndStatex Entropy  rate $\beta > 0$ (default $0.1$),
\algrule
\State Initialize $h, v$ as described in \secref{init}
\State $\forall t\in [N],  \bu_t \gets \mzero_K$
\Do{}
    \IndStatex // [E-step]
    \IndStatex // Compute the allocation $\bgamma_t \in \Delta^{K-1}$
    \For{$t = 1,...,N$}
        \InddStatex // $\odot$ denotes element-wise multiplication
        \State $\bgamma_t \gets \expbnf{\squared{y_t\vone_K - h(\bx_t)}}{2\sigma^2}   \odot \func{\softmax}{\bu_t + v(\bx_t)}$
        \State $\bgamma_t \gets \bgamma_t / (\bgamma_t^\T \vone_K)$
    \EndFor
    \IndStatex // [M-step]
    \State Update $h, v,$ and $\{\bu_t \}_{t=1}^N$ using the Adam optimizer~\cite{adam2015kigma} with learning rate $\eta$ based on $L(h,v,\{\bu_t\}; \{\gamma_t\})$ (\eqref{Ldef})
\Until{Convergence}
\algrule
\Ensure Prediction model $h:\scriptX\to\scriptY^K$,
\IndStatex Mapping function $v:\scriptX \to \doubleR^{K}$,
\IndStatex Weights $\{\bu_t\}_{t=1}^{N} \in \doubleR^{N \times K}$
\end{algorithmic}
\end{algorithm}

\subsubsection{E-Step}
In E-step, we maximize $F(q)$ w.r.t. $q$.
Note that the left hand in \eqref{em} does not depend on $q$ and hence maximizing $F(q)$ is equivalent to minimize $\DKL{q(\bz_t)}{ p(\bz_t|\bx_t, y_t, h,v,\bu_t)}$, which is naively achieved by letting $q(\bz_t)$ be $p(\bz_t|\bx_t, y_t, h,v,\bu_t)$, making the KL term to be 0.
Then, we compute and fix the probability allocation $\gamma_{tk} := q(\bz_t = \be_k)$ for every $k \in [K]$ (lines 4 to 7 in \algref{scd}) for M-step;
By Bayes's theorem, we have
\begin{align}
    q(\bz_t) &= p(\bz_t \mid \bx_t, y_t, h, v, \bu_t) \\
    &= \frac{p(y_t, \bz_t \mid \bx_t, h, v, \bu_t)}{\sum_{\bz_t' \in \doubleB^K}p(y_t, \bz_t' \mid \bx_t, h, v, \bu_t)}. \eqlab{b1}
\end{align}
Then, by the dependencies depicted in \figref{graphicalmodel}, we have
\begin{align}
    &p(y_t, \bz_t = \be_k \mid \bx_t, h, v, \bu_t) \notag \\
    &= \frac{1}{\sqrt{2\pi \sigma^2}} \expbnf{\squared{y_t - h_k(\bx_t)}}{2\sigma^2}  \func{\softmax}{\bu_t + v(\bx_t)}_k.  \eqlab{b2}
\end{align}
Combining \eqref{b1} and \eqref{b2}, we obtain the allocation vector $\bgamma_t:= [\gamma_{t1},...,\gamma_{tK}]^\T \in \Delta^{K-1}$ for the current iteration.
The computation of $\bgamma_t$ is repeated for all $t \in [N]$.

\subsubsection{M-Step}
In M-step, we maximize $F(q)$ w.r.t. $h,v$, and $\bu_t$.
Note that the previous E-step has set the KL term to 0.
This means that the left hand of \eqref{em}, $\log p(y_t|\bx_t, h,v, \bu_t)$ increases as long as $F(q)$ increases.
Since $q(\bz_t = \be_k) = \gamma_{tk}$, we have
\begin{align}
    &F(q) = \sum_{\bz_t \in \doubleB^K} q(\bz_t) \log \frac{p(y_t,\bz_t \mid \bx_t, h,v, \bu_t)}{q(\bz_t)}\\
    &= - \frac{\bgamma_t^\T \squared{y_t\vone_K  - h(\bx_t)}}{2\sigma^2} + H(\bgamma_t,\func{\softmax}{\bu_t + v(\bx_t)})  + C_F,
\end{align}
where $\vone_K := [1,...,1]^\T \in \doubleR^K$ is the vector of ones,
$C_F := - \sum_{k=1}^K \gamma_{tk} \left( \log {\sqrt{2\pi\sigma^2}} + \log \gamma_{tk} \right)$ is independent from $h, v,$ and $\bu_t$,
and $H(p,q) := \sum_{\bz \in \doubleB^K} p(\bz) \log q(\bz) $ compute the cross-entropy between $p$ and $q$.
Since $C_F$ can be safely ignored when optimizing $h,v$, and $\{\bu_t\}$, the overall loss, which we aim to minimize by Adam~\cite{adam2015kigma} in M-step, is defined as
\begin{align}
    &L(h,v,\{\bu_t\}; \{\bgamma_t\}) \notag \\
    &:= \sum_{t=1}^N \frac{\bgamma_t^\T \squared{y_t\vone_K  - h(\bx_t)}}{2\sigma^2} - H(\bgamma_t,\func{\softmax}{\bu_t + v(\bx_t)})  \notag \\
    &\qquad + \alpha \sum_{t=1}^{N} \enorm{\bu_t - \bu_{t+1}}^2 
    + \beta \sum_{t=1}^{N} H_E(\func{\softmax}{\bu_t + v(\bx_t)})
    ,\eqlab{Ldef}
\end{align}
where we add a smoother term w.r.t. the change of $\bu_t$ and a regularization term w.r.t. the number of active source components with hyperparameters $\alpha$ and $\beta > 0$.
$H_E(p) := H(p,p)$ is defined as the entropy of $p$ and we abuse to define $\bu_{N+1} := \vzero_K$, i.e., the vector of zeros.
We iteratively conduct E-step and M-step until convergence to obtain $h,v$, and $\{\bu_t\}_{t=1}^N$.



\subsubsection{Initialization of $h$ and $v$}
\seclab{init}
EM algorithms are often sensitive to initialization.
To address this, we use a heuristic approach that delivers satisfactory results.
First, we initialize $h$ to fit $D$ with diverse outputs, ensuring each $h(\bx)_k$ for $k \in [K]$ handle a different source component.
The initial $h$ is set as
\begin{align}
    h = \argmin_{h\in \scriptH} \sum_{(\bx,y) \in D,k \in [K]}\squared{h(\bx_i)_k - y_i +  \frac{3\sigma(2k-K-1)}{K-1}}. 
\end{align}
Next, we initialize $v$, which includes $\bc_k$ and $\bs_k \in \doubleR^d$, by sampling $\bc_k \sim \ndist{\vzero_d}{\mI_d}$ and setting $\bs_k = \vone_d$ for each $k \in [K]$, where $\ndist{\vzero_d}{\mI_d}$ denotes the $d$-dimensional Gaussian distribution.

\subsubsection{Selection of $K$ of \algref{scd}}
\seclab{Ktune}

\begin{algorithm}[t]
\caption{Source component decomposition with automated data-driven tuning of $K$.}
\alglab{sks}
\begin{algorithmic}[1]
\Require Training data $D = \{(\bx_t, y_t)\}_{t=1}^N$
\algrule
\State Split $D$ into $\Dtr$ and $\Dte$ as described in \secref{Ktune}.
\State $h' \gets \argmin_{h \in \scriptH'} \sum_{(\bx,y) \in \Dtr} \squared{h(\bx) - y} $
\State $\xi \gets \sqrt{1/ |\Dva|\sum_{(\bx,y) \in \Dva} \squared{h'(\bx) - y}}$
\For{$K'=2$ to $10$}
    \State $\sigma \gets \xi/\sqrt{K'} $
    \State $(h_{K'},v_{K'},\{\bu_t\}) \gets$  Run \algref{scd} with $(\Dtr, K', \sigma)$
    \State $LL_{K'} \gets \log \mathrm{likelihood}(h_{K'}, v_{K'}, \{\bu_t\}; \Dva)$ with \eqref{LLdef}
\EndFor
\State $K \gets \argmax_{{K'} \in \{2,...,10\}} LL_{K'} $
\algrule
\Ensure Prediction model $h_K:\scriptX\to\scriptY^K$,
\IndStatex Mapping function $v_K:\scriptX \to \doubleR^{K}$
\end{algorithmic}
\end{algorithm}

We need to determine $K$ and $\sigma$ to run \algref{scd}.
For this purpose, we employ a data-driven approach.
We first split the entire dataset $D$ into training data $\Dtr$ and validation data $\Dva$ in the following manner;
We re-index each sample of $D$ by $t=1, \ldots, m-1, m, m, m+1, \ldots, 2m-1, 2m, 2m, 2m+1, \ldots$ with, e.g., $m=4$.
Then, we use samples with duplicated indices for $\Dva$, and the rest for $\Dtr$, keeping the continuity of $t$ in $\Dtr$ while splitting out $\Dva$ for evaluation.


The base quantity $\xi > 0$ used to determine $\sigma$ is computed as the validation error of an offline model $h':\scriptX \to \scriptY$ as
\begin{align}
    \xi  := \sqrt{\frac{1}{|\Dva|} \sum_{(\bx,y) \in \Dva} \squared{h'(\bx) - y}}, \eqlab{xi}
\end{align}
where $h' = \argmin_{h \in \scriptH'} \sum_{(\bx,y) \in \Dtr} \squared{h(\bx) - y} $ is trained with $\Dtr$ with an arbitrary model space $\scriptH'$.
Among each candidate of $K' \in \{2,...,10\}$,\footnote{The range of $K$ follows that of FUZZ-CARE~\cite{fuzz2020song}, however, a wider range can be used in practice.} we run \algref{scd} with $\sigma = \xi / \sqrt{K'}$,\footnote{$\sigma = \xi / \sqrt{K'}$ is employed based on empirical experiments.} and evaluate the log likelihood using $\Dva$ as
\begin{align}
    \log \mathrm{likelihood}(h, v, \{\bu_t\}; \Dva) := \sum_{i=1}^{|\Dva|} \log p(y_{i}| \bx_{i}, h, v, \bu_{mi}) \eqlab{LLdef}.
\end{align}
Finally, we select $K$ which achieves the highest log likelihood among all candidates of $K$, outputting trained $h$ and $v$ with selected $K$.
These $h$ and $v$ are then used for mixing weight adaption, elaborated in the next section.



\subsection{Mixing Weight Adaptation}
\seclab{MWA}

Now, the task is to adaptively ensemble the output of $h$ with $v$ by optimizing $\bu_t$ from $t=N+1$ until $N+T$, as defined in \defref{mwa}.
In this section, we abuse the notation to re-index $t = N+1,..., N+T$ to be $t = 1,...,T$ for better readability.
We use  $L_t(\{\bu_t\})$ to denote the cumulative loss at $t$ as 
\begin{align}
    L_t(\{\bu_t\})
    &:= \sum_{s \in [t]} \ell_s(\bu_s) \\
    \ell_t(\bu) &:= \squared{ \func{\softmax}{\bu + v(\bx_t)}^\T h(\bx_t) - y_t }
\end{align}
for every $t \in [T]$ and we let $\bu_1$ be $\vzero_K$.

\begin{algorithm}[t]
\caption{Mixing weight adaptation (based on Algorithm 2~\cite{oogd2023}).}
\alglab{mra}
\begin{algorithmic}[1]
\Require Prediction model $h:\scriptX \to \scriptY^K$,
\IndStatex Mapping function $v: \scriptX \to \doubleR^K$
\Parameters Number of base algorithms $M \in \doubleZ_{\geq 2}$ (default $11$),
\IndStatex Minimum learning rate of base-learner $\eta > 0$ (default $0.01$),
\IndStatex Learning rate of meta-learner $\varepsilon > 0$ (default 1.0),
\IndStatex Correction coefficient $\lambda > 0$ (default $0.1$)
\algrule
\State $\forall i \in [M], \bu_{1i}, \bu'_{1i} \gets \vzero_{K},  \eta_i \gets 2^{i-1}\eta$
\State $\bp_1 \gets \vone_{M}/M$
\For{$t = 1 ,..., T$}
    \IndStatex // Predict
    \State Observe $\bx_t \in \scriptX$
    \State $\bu_t \gets [\bu_{t1},...,\bu_{tM}] \bp_t$
    \State $\hat{y}_t \gets  \func{\softmax}{\bu_t + v(\bx_t)}^\T h(\bx_t)$
    \State Observe $y_t \in \scriptY$ and suffer the loss $\squared{\hat{y}_t - y_t}$
    \IndStatex // Update
    \State $\bell_t \gets \vzero_K$, $\bm{m}_{t+1} \gets \vzero_K$
    \For{$i=1,...,M$}
        \State $\bu'_{(t+1)i} \gets \bu'_{ti} - \eta_{i} \nabla \ell_t(\bu_t)$
        \State $\bu_{(t+1)i} \gets \bu'_{(t+1)i} - \eta_{i} \nabla \ell_t(\bu_t)$
        \If{$t \geq 2$}
            \State $\ell_{ti} \gets  \innerprod{\nabla \ell_t(\bu_t)}{ \bu_{ti}} + \lambda \enorm{\bu_{ti} - \bu_{(t-1)i}}^2 $
            \State $m_{(t+1)i}$
            \IndddStatex \quad$\gets  \innerprod{\nabla \ell_t(\bu_t)}{ \bu_{(t+1)i} } + \lambda \enorm{\bu_{(t+1)i} - \bu_{ti}}^2$
        \Else
            \State $\ell_{1i} \gets  \innerprod{\nabla \ell_1(\bu_1)}{ \bu_{1i}}$, $m_{2i} \gets \lambda \enorm{\bu_{2i} - \bu_{1i}}^2$
        \EndIf
    \EndFor
    \State $\bp_{t+1} \gets \func{\softmax}{ - \varepsilon  \left(\bm{m}_{t+1} + \sum_{s=1}^t  \bell_s  \right) }$
\EndFor
\algrule
\Ensure Cumulative loss $\sum_{t=1}^T \squared{\hat{y} - y_t}$
\end{algorithmic}
\end{algorithm}

The minimization of $L_t(\{\bu_t\})$ can be regarded as an online convex optimization (OCO) w.r.t. $\bu_t$.
As a representative approach to OCO, we employ a two-layer dynamic regret minimization algorithm~\cite[Algorithm 2]{oogd2023}, as fully described in \algref{mra}.
The bottom (base-) layer consists of $M \in \doubleZ_{\geq 2}$ optimistic online gradient descent (OGD) algorithms which provide $\bu_{ti} \in \doubleR^K$ for each $i \in [M]$.
Each base-learner utilizes an internal variable $\bu'_{ti} \in \doubleR^K $ and updates it for better estimation of $\bu_{ti}$ as
\begin{align}
    \bu'_{(t+1)i} &= \bu'_{ti} - \eta_{i}  \nabla \ell_t(\bu_{ti}) \\
    \bu_{(t+1)i} &= \bu'_{(t+1)i} - \eta_{i}  \nabla \ell_t(\bu_{ti}),
\end{align}
where $\eta_i >0$ is the learning rate and we set $\bu'_{1i} := \vzero_K$.

The top (meta-) layer is realized by an optimistic hedge algorithm, which dynamically ensembles the outputs by the base-learners.
Both base- and meta-learners update their internal variables based on the gradient $\nabla \ell_t(\bu_t)$ as computed as 
\begin{align}
    \nabla \ell_t(\bu_t)
    &:= \frac{\partial}{\partial \bu_t} \squared{ \func{\softmax}{\bu_t + v(\bx_t)}^\T h(\bx_t) - y_t } \\
    &= 2\frac{\ba_t \odot \bb_t}{ \ba_t^\T  \bb_t} \odot \left( 
    h(\bx_t) - \frac{ \ba_t^\T (h(\bx_t) \odot \bb_t)}
        { \ba_t^\T  \bb_t } \vone_K
    \right) \notag \\ &\qquad\qquad \times
    \left(\frac{\ba_t^\T (h(\bx_t) \odot \bb_t)}{ \ba_t^\T \bb_t } - y_t \right),
\end{align}
where we define $\ba_t, \bb_t \in \Delta^{K-1}$ by $\ba_t := \softmax(\bu_t)$ and $\bb_t := \softmax(v(\bx_t))$.

Regarding the hyperparameters including the number of the base algorithms $M \in \doubleZ_{\geq 2}$, the smallest learning rate of base algorithms $\eta > 0$, the learning rate of the meta-learner $\varepsilon >0$, and the correction coefficient $\lambda >0$, we employ to set default parameters based on empirical experiments as shown in \algref{mra}.
Note that although the authors of the paper propose theoretical calculation of the hyperparameters that guarantees the worst case performance, these hyperparameters are rather conservative, and our setting based on empirical experiments often demonstrate better performance in practice.

\section{Experiments}
We conduct numerical experiments using several real-world-based s to verify the effectiveness of our method.
The results are summarized as follows.
\begin{slimitemize}
\item Our method (SCSDA) achieves the best or at least comparable results on five out of six datasets, reducing cumulative loss by up to 67.4\%.
\item While baseline methods show variable performance across datasets, our method consistently delivers superior results.
\item Our method excels in accurately decomposing source components in offline settings, demonstrating that our sequential decomposition and adaptation approach is promising.
\end{slimitemize}

\subsection{Dataset}
The dataset statistics are summarized in \tabref{dataset}.
Solar dataset is obtained from Kaggle data repository, while others are obtained from UCI Machine Learning Repository.
As pre-process, the input and output variables are normalized to have zero mean and unit variance and we drop samples with NA/nan values.

\begin{table}[t]
\caption{Dataset statistics.}
\label{tab:dataset}
\centering
\setlength\tabcolsep{13pt}
\begin{tabular}{lrrl} \toprule
Dataset   &Samples&Features&Data source\\ \midrule
Seoul bike&8760   &15      & UCI~\cite{seoulbike2020,seoulbike}   \\
Bike sharing&17379  &9       & UCI~\cite{bikesharing2013Fanaee-T,bikesharing}  \\
Solar      &32686  &5       &  Kaggle~\cite{solar}    \\
Birmingham &35501  &33      &  UCI~\cite{birmingham}    \\
PM 2.5       &41757  &11      &  UCI~\cite{pm25xuan,pm25}  \\
Tetouan     &52416  &7       &  UCI~\cite{tetouan2017,tetouan}  \\
\bottomrule
\end{tabular}
\end{table}

\begin{table*}[t]
\caption{Average cumulative squared error $\widehat{R}(\{f_t\}; \Dte)$ over 30 random trials ($\downarrow$), comparing our SCSDA with existing methods.
Numbers after $\pm$ represent standard deviations.
$\textbf{Boldfaces w/ star}^*$ highlight the lowest errors and basic \textbf{boldfaces} show comparable results based on the Wilcoxon signed-rank test~\cite{wilcoxon1945} with the significance level of $5\%$.
Gain shows the gain of our SCSDA compared to the best baseline result if the difference is significant.
}
\label{tab:result}
\centering
\setlength\tabcolsep{3pt}
\begin{tabular}{c|cccccc|r} \toprule
Dataset&Offline&OGD&SEGA~\cite{song22sega} &Condor~\cite{zhao18condor}&FUZZ-CARE~\cite{fuzz2020song} &SCSDA (Ours) & Gain [\%] \\ \midrule
Seoul bike &$375.6\pm115.7$ &$\bhm{307.8\pm56.4}$&$2257.4\pm1461.9$&$497.7\pm127.6$ &$367.0\pm81.3$ &$\chm{318.6\pm126.8}$ &($\approx 0.00$)\\ 
Bike sharing&$443.7\pm181.1$ &$421.9\pm169.5$ &$5326.6\pm5687.3$&$660.3\pm370.9$ &$539.6\pm235.9$ &$\bhm{137.6\pm59.2}$&$-67.39$\\
Solar &$327.6\pm97.5$ &$192.2\pm60.7$ &$1868.9\pm1764.2$&$\chm{160.0\pm52.7}$&$\bhm{155.6\pm46.6}$ &$\chm{167.0\pm77.0}$&($\approx 0.00$)\\ 
Birmingham &$310.10\pm111.11$&$242.76\pm81.52$ &$1291.61\pm513.85$ &$\bhm{132.28\pm29.32}$&$226.28\pm86.45$ &$159.37\pm55.07$ & $+20.48$ \\
PM 2.5 &$629.0\pm223.7$ &$424.7\pm131.3$ &$6639.7\pm5999.8$&$429.9\pm199.6$ &$\bhm{354.5\pm116.5}$&$\chm{396.5\pm167.9}$&($\approx 0.00$)\\ 
Tetouan &$173.0\pm56.4$ &$119.8\pm41.1$ &$3278.6\pm2601.1$&$159.9\pm41.5$ &$168.5\pm33.9$ &$\bhm{51.4\pm29.6}$&$-57.10$\\
\bottomrule
\end{tabular}
\end{table*}

\subsection{Comparison Methods}
We use the following baselines and our method in the experiments.
To ensure a fair comparison, all methods employ a neural network (NN) with the same architecture and training strategy.
Specifically, we use a 3-layer multilayer perceptron (MLP) with 128 hidden units, applying the Swish activation function~\cite{swish2017}.
The parameters are optimized using full-batch training with the Adam optimizer~\cite{adam2015kigma}, a learning rate of 0.01, and a weight decay of $1 \times 10^{-4}$, over 200 epochs.
We select the parameters that achieve the lowest validation error as the final model parameters, using a randomly chosen 10\% of the training data for validation.
\begin{slimitemize}
\item Offline: A simple baseline where a vanilla NN model is trained using $D$ and predicts $\Dte$ without any adaptation.
\item OGD: Online Gradient Descent.
A NN is initialized with $D$.
For $t \geq N+1$, the model is updated by gradient descent using $(\bx_t, y_t)$.
We set the learning rate to be $1\times 10^{-3}$, which yields the best average results for the experiment.
\item SEGA~\cite{song22sega}: A model-pool based method for regression.
SEGA processes each sample of $D$ and $\Dte$.
It trains a new model every 100 samples and adds it to the pool.
SEGA selects the best model in the pool for predictions using the idea of \textit{drift gradient}.
Default hyperparameters are used.
\item Condor~\cite{zhao18condor}: A model-pool based method.\footnote{Condor is initially proposed for classification, however, it can handle regression tasks simply by replacing its loss function to the squared error.}      Similar to SEGA, Condor processes each sample of $D$ and $\Dte$.
It trains a new model and expands the pool either every 100 samples or when data drift is detected, reusing the parameter ensemble of models in the pool.
Default hyperparameters are used.
\item FUZZ-CARE~\cite{fuzz2020song}: A model-pool based method for regression.
Similar to our method, FUZZ-CARE decomposes and learns multiple prediction models, and then ensembles them for prediction.
It updates the parameters online for every new sample, and in addition, every 100 samples observed, it decomposes and learns completely new prediction models.
We adjust the algorithm for using NN, employing gradient decent and set the learning rate to be 0.01 for best performance.
\item SCSDA\footnote{SCSDA is short for Source Component Shift adaptation via Decomposition and Adaptation.} (Ours): Our proposed method, detailed in \secref{method}.
It decomposes multiple source components using $D$, followed by online adaptation of mixing weights.
Default parameters are used.
\end{slimitemize}

\subsection{Setting}
\begin{figure}[t]
\centering
\includegraphics[width=0.85\linewidth]{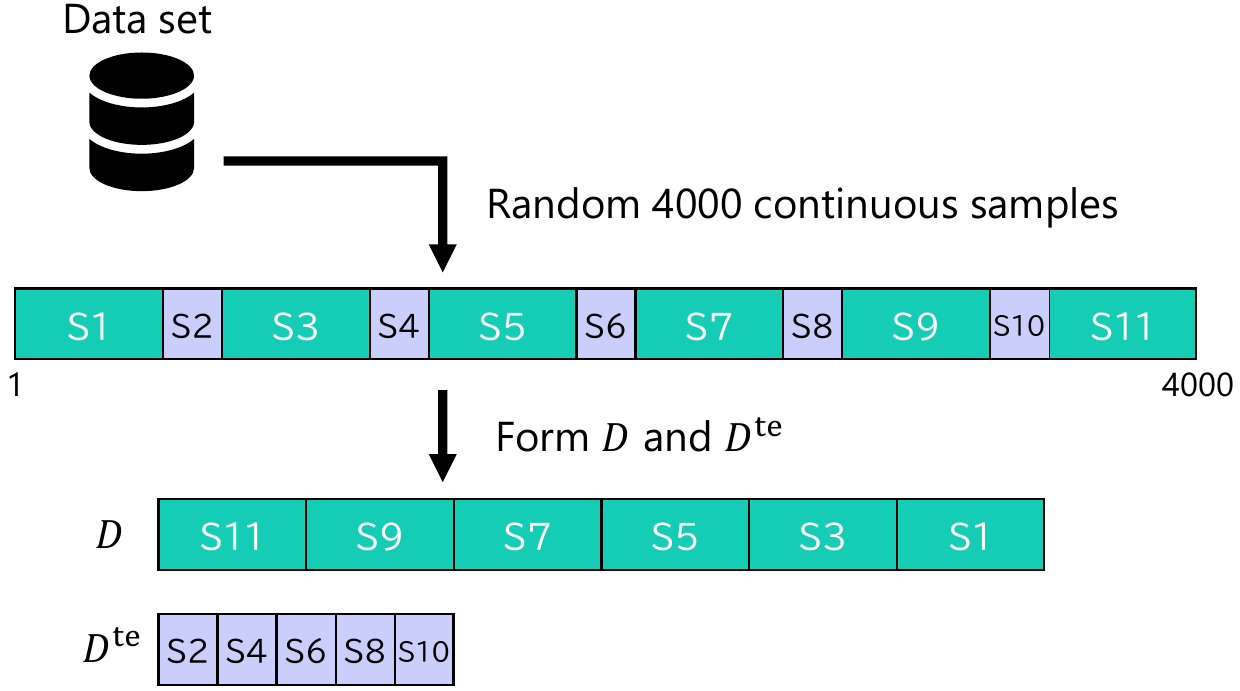}
\caption{Preparation of $D$ and $\Dte$, simulating source component shifts.}
\label{fig:eurai}
\end{figure}

\noindent
To simulate source component shifts, we prepare $D$ and $\Dte$ by selecting 4000 continuous samples from a dataset, starting at a random index.
Out of these, samples indexed from 501 to 700, 1201 to 1400, 1900 to 2100, 2601 to 2800, and 3301 to 3500 are set aside for $\Dte$, with the remainder used for $D$.
The samples are arranged in order as shown in \figref{eurai} so that $D$ and $\Dte$ are continuous.
Even if the original dataset lacks source component shifts, this configuration simulates such shifts.

We run each comparison method to predict each sample of $\Dte$ using $D$ and evaluate the cumulative loss $\widehat{R}(\{f_t\}; \Dte)$ of \eqref{scsap}.
Each experiment is repeated 30 times with different random seeds, and we report the average loss along with standard deviations.

\subsection{Result}
The results are shown in \tabref{result}.
Our method achieves the best or comparable results in five out of six datasets, demonstrating its effectiveness for source component shift adaptation.
Notably, it reduces cumulative losses by 67.4\% and 57.1\% for the Bike Sharing and Tetouan datasets, respectively.
While Condor outperforms our method for Birmingham dataset, our method shows the second best result, beating the other baselines.

Among the baselines, SEGA generally fails to achieve accurate predictions.
The performance of Condor and FUZZ-CARE depend on the datasets and are often worse than the results of offline, which does not employ any adaptation mechanisms.

These results empirically demonstrate the superiority of our method for source component shift adaptation over the baselines.

\subsection{Decomposability of Source Components}
\seclab{offline}

We further assess the quality of the model pools prepared before predicting samples of $\Dte$.
The experimental setup remains consistent with previous experiments, while we disallow baseline methods to add new models or update models in the pool.
Instead, these methods can only prepare models offline using $D$ and modify ensemble weights online for $\Dte$ if they use an ensemble approach.
Note that by design, our method does not add or update models during adaptation phases.

The results are summarized in \tabref{offline}.
Our method achieves the best results across almost all datasets.
Condor performs worse than in previous experiments, highlighting its insufficient model pool preparation for adapting to source component shift.
In contrast, FUZZ-CARE shows limited performance degradation compared to Condor, due to its effective data decomposition and model pool management.
However, its performance is inferior to our SCSDA.
These results demonstrate the superior and accurate decomposability of source components of our method, showcasing our sequential decomposition and adaptation approach is valid and promising.

\begin{table}[t]
\caption{Average cumulative squared error $\widehat{R}(\{f_t\}; \Dte)$ over 30 random trials ($\downarrow$), comparing performance w/o model addition and updates.}
\label{tab:offline}
\centering
\setlength\tabcolsep{3.5pt}
\begin{tabular}{c|cccc} \toprule
\multirow{3}{*}{Dataset}&SEGA w/o &Condor w/o &\multirow{2}{*}{FUZZ-CARE w/o}&\multirow{2}{*}{SCSDA} \\ 
&model add.&model add.&\multirow{2}{*}{model updates}&\multirow{2}{*}{(Ours)} \\ 
&for $\Dte$&for $\Dte$&&\\ 
\midrule
\multirow{2}{*}{Seoul bike} &$2560.6$ &$818.8$ &$372.5$ &$\bhm{318.6}$ \\
 &$\pm1981.4~~$ &$\pm310.0~~$ &$\pm83.9~~$ &$\pm\chm{126.8}~~$\\
[0.3em]
\multirow{2}{*}{Bike sharing}&$3323.1$ &$920.4$ &$516.2$ &$\bhm{137.6}$ \\
 &$\pm4699.3~~$ &$\pm656.1~~$ &$\pm226.8~~$ &$\pm\chm{59.2}~~$ \\
[0.3em]
\multirow{2}{*}{Solar} &$1671.3$ &$468.9$ &$\chm{173.1}$ &$\bhm{167.0}$ \\
 &$\pm2310.2~~$ &$\pm130.0~~$ &$\pm\chm{43.1}~~$ &$\pm\chm{77.0}~~$ \\
[0.3em]
\multirow{2}{*}{Birmingham}&$1357.91$&$586.90$&$240.75$&$\bhm{159.37}$ \\
&$\pm753.98~~$&$\pm129.95~~$&$\pm89.41~~$&$\chm{\pm55.07~~}$ \\
[0.3em]
\multirow{2}{*}{PM 2.5} &$7714.1$ &$868.2$ &$\bhm{396.2}$ &$\chm{396.5}$ \\
 &$\pm11183.7~~$ &$\pm373.1~~$ &$\pm\chm{129.1}~~$ &$\pm\chm{167.9}~~$\\
[0.3em]
\multirow{2}{*}{Tetouan} &$4315.6$ &$510.0$ &$190.5$ &$\bhm{51.4}$ \\
 &$\pm6578.8~~$ &$\pm185.1~~$ &$\pm41.1~~$ &$\pm\chm{29.6}~~$ \\
\bottomrule
\end{tabular}
\end{table}

\section{Conclusion}
This paper dealt with the source component shift adaptation problem in regression tasks.
We theoretically divided the source component shift adaptation problem into two distinct subproblems: offline source component decomposition and online mixing weight adaptation.
On this basis, we developed a source component shift adaptation method via an offline decomposition and online mixing approach, leading to enhanced prediction performance.
Experiments on multiple real-world-based regression datasets highlighted the superiority of our method over the baseline methods.

While our method is specifically designed for source component shifts, its theoretical validity and empirical effectiveness in cases where source component shifts and other types of data shifts are mixed remain to be seen.
In addition, \assref{nonzero}, which is essential for our analysis, may be relaxed to better suit real-world applications.
Exploring these issues as future work would further broaden the applicability of our proposed method.





\bibliography{reference}

\supplementary

\section*{Supplementary Material}

\section{Dataset Description}

\block{Seoul bike~\cite{seoulbike2020,seoulbike}.}
This dataset is introduced in~\cite{seoulbike2020} and obtained from UCI machine learning data repository~\cite{seoulbike}.
This contains 8760 samples of hourly count of rental bikes from 00:00 on December 2, 2017 to 23:00 on November 30, 2018 with meteorological information.
The input features are hour, temperature, humidity, wind speed, visibility, dew point temperature, solar radiation, rainfall, snowfall, holiday or not, functioning day  or not, and one-hot encoding of seasons (spring, summer, autumn, winter), totaling in fifteen features. 
The objective is to predict the number of bikes rented per hour.

\block{Bike sharing~\cite{bikesharing2013Fanaee-T,bikesharing}.}
This dataset is introduced in~\cite{bikesharing2013Fanaee-T} and obtained from UCI machine learning data repository~\cite{bikesharing}.
This contains 17379 samples of hourly count of rental bikes from 00:00 on January 1, 2011 to 23:00 on December 31, 2012 in Capital bikeshare system in Washington, DC with the corresponding weather and season information.
The nine input features we used are hour, holiday or not, weekday or not, working day or not, weather situation (clear, mist, light rain/snow, or, heavy rain/snow), temperature in Celsius, feeling temperature in Celsius, humidity, and wind speed.
The target variable is the count of total rental bikes including both casual and registered uses.

\block{Solar~\cite{solar}.}
This dataset is obtained from Kaggle Dataset~\cite{solar}.
The description is detailed in~\cite{fuzz2020song} as follows:
``\textit{Solar is provided by NASA and contains 32686 records of meteorological data from the hawaii space exploration analog and simulation (HI-SEAS) weather station from 23:55:26 on September 29, 2016 to 00:00:02 on December 1, 2016 (Hawaii time) in the period between Missions IV and V. The data interval is roughly 5 min. The input features are temperature (unit: degrees Fahrenheit), humidity (unit: percent), barometric pressure (unit: Hg), wind direction (unit: degree), wind speed (unit: miles per hour), and the target variable is solar radiation (unit: watts per $\text{meter}^2$).}''

\block{Birmingham (Parking Birmingham)~\cite{birmingham}.}
The data is obtained from UCI machine learning data repository~\cite{birmingham}.
This is collected from car parks in Birmingham, UK, operated by NCP and Birmingham City Council, from 07:59:42 on October 4, 2016, to 16:30:35 on December 19, 2016, containing a total of 35501 samples.
Each sample is collected for each car park for roughly every 30 minutes.
The input features we used are one-hot encoding of 31 car parks, hour, minute, and the capacity of the park code, totaling in 33 features.
The target is the occupation ratio of the car park.

\block{PM 2.5 (Beijing PM2.5)~\cite{pm25xuan,pm25}.}
The data is introduced in~\cite{pm25xuan} and obtained from UCI machine learning data repository~\cite{pm25}.
This data contains 41757 samples of hourly PM2.5 data and corresponding meteorological data at US Embassy in Beijing, China from 00:00 on January 1, 2010 to 23:00 on December 31, 2014.
The input features are
hour, dew point, temperature, air pressure, one-hot encoding of combined wind direction (NE, NW, SE, cv), cumulated wind speed, cumulated hours of snow, and  cumulated hours of rain, consisting of 11 features.
The target is PM2.5 concentration (ug/$\text{m}^3$).

\block{Tetouan (Power Consumption of Tetouan City)~\cite{tetouan2017,tetouan}.}
This data is introduced in~\cite{tetouan2017} and obtained from UCI machine learning data repository~\cite{tetouan}.
This contains 52416 records of power consumption for three zones in Tetouan city, recorded every 10 minutes from 00:00 on January 1, 2017, to 23:50 on December 30, 2017.
The input features are hour, minute, temperature, humidity, wind speed, general diffuse flows, and diffuse flows.
The target we used is the sum of the power consumption in zone 1, 2, and 3.

\section{Running Time of Methods}
We compare the running times of the methods used in the experiments in Section 4.
Although we use the Tetouan dataset~\cite{tetouan2017,tetouan} as a representative example, the differences in running times across different datasets are minimal.

\begin{figure}[h]
\centering
\includegraphics[width=0.5\linewidth]{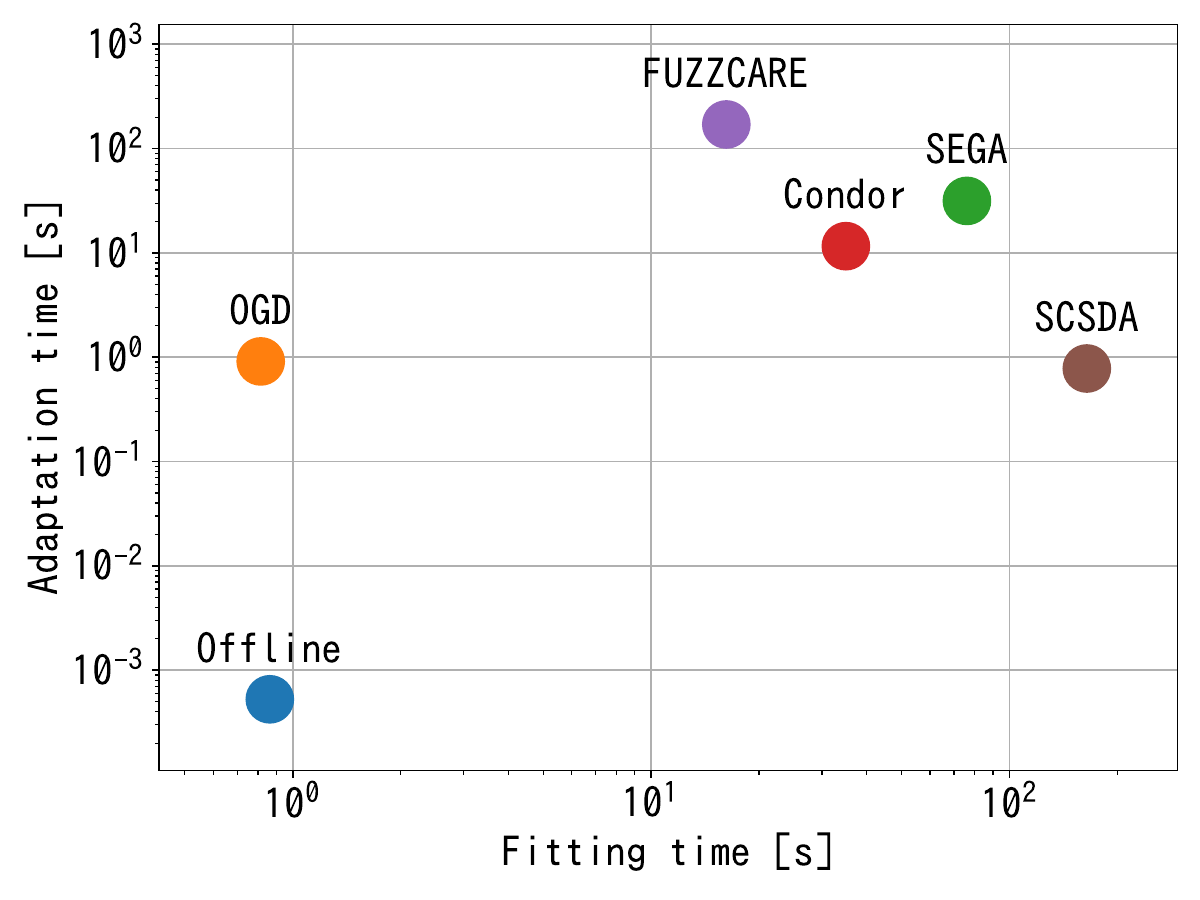}
\caption{Average running time (seconds) for Tetouan dataset over 30 trials.}
\label{fig:runtime}
\end{figure}

The running time results are reported in \figref{runtime}.
The horizontal axis represents fitting time: training time for Offline and OGD, time to run SEGA and Condor on $D$, and time for decomposition and learning prediction models for FUZZ-CARE and SCSDA (ours).
The vertical axis indicates adaptation time: prediction time for Offline, and prediction, model update, and model addition time for other methods.
Offline, which trains a model on $D$ and predicts $\Dte$ without any adaptation, is unsurprisingly the fastest in fitting and adaptation time.
Though OGD is as fast as Offline in fitting, it requires more time for adaptation due to gradient descent updates.
Our SCSDA takes the longest time for fitting due to its detailed source component decomposition.
However, it is the second-fastest for adaptation, thanks to the numerically efficient adaptation algorithm described in Section 3.3.

Other baseline methods, such as FUZZ-CARE, Condor, and SEGA, require less fitting time but more adaptation time than our method.
Hence, these methods are concluded to be less effective, offering inferior predictive performance and higher computational costs for adaptation.

Overall, these results highlight the efficiency of our method in computational adaptation costs.

\section{Additional Experiments on Raw Real-World Datasets}
The experiments in Section 4 are based on simulated source component shifts.
To assess our method's effectiveness with raw drifting datasets, where source component shifts and other types of data shifts might occur simultaneously, we conducted additional experiments.

\begin{figure}[h]
\centering
\includegraphics[width=0.4\linewidth]{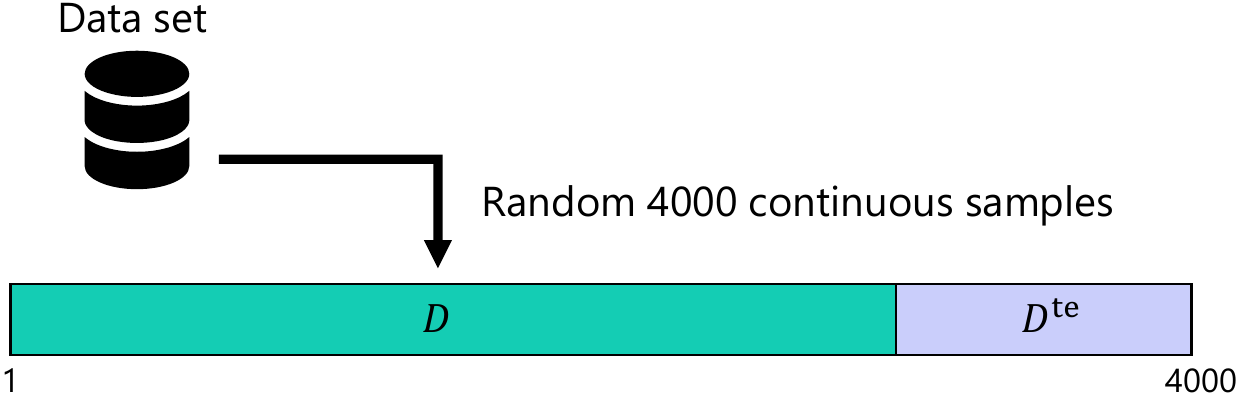}
\caption{Preparation of $D$ and $\Dte$, simulating source component shifts.}
\label{fig:additional}
\end{figure}

For the preparation of $D$ and $\Dte$, we randomly select 4000 consecutive samples.
The first 3000 samples are used for $D$, and the subsequent 1000 samples are used for $\Dte$, as shown in \figref{additional}.
Since the source component shifts are not ensured for this setting, our method, specialized for the shifts, may fail for accurate prediction.
Other settings remain the same as in previous experiments.

The results are displayed in \tabref{stream}, which is related to \tabref{result}.
We also evaluated offline source component decomposition, detailed in \secref{offline}, and reported in \tabref{offlinestream}.
As shown, our method no longer performs best and OGD, Condor, and FUZZ-CARE, which add new models and/or update models online, perform well.
Yet, our method still excels or performs comparably for Bike sharing and Tetouan datasets.
This suggests our method can outperform baselines without strict source component shift controls.

Moreover, \tabref{offlinestream} shows that our method performs best in four out of six datasets in offline settings.
This demonstrates its superiority in delivering better prediction models over baselines, even without source component shifts being maintained.

Overall, our method is confirmed to be effective even without controlled source component shift simulation.

\begin{table}[h]
\caption{Average cumulative squared error $\widehat{R}(\{f_t\}; \Dte)$ over 30 random trials ($\downarrow$) with datasetting shown in \figref{additional}.}
\label{tab:stream}
\centering
\setlength\tabcolsep{8pt}
\begin{tabular}{c|cccccc|r} \toprule
Dataset&Offline&OGD&SEGA~\cite{song22sega} &Condor~\cite{zhao18condor}&FUZZ-CARE~\cite{fuzz2020song} &SCSDA (Ours) & Gain [\%] \\ \midrule
\multirow{2}{*}{Seoul bike}&$1507.32$&$\bhm{388.43}$&$2786.59$&$562.11$&$429.57$&$1233.62$ & \multirow{2}{*}{$+217.59$} \\
&$\pm1376.41~~$&$\chm{\pm72.07~~}$&$\pm1375.44~~$&$\pm173.46~~$&$\pm73.12~~$&$\pm1309.01~~$ &   \\
[0.3em]
\multirow{2}{*}{Bike sharing}&$592.76$&$498.07$&$3543.84$&$762.90$&$620.09$&$\bhm{219.59}$  & \multirow{2}{*}{$-55.91$}\\
&$\pm258.11~~$&$\pm194.57~~$&$\pm3476.78~~$&$\pm434.18~~$&$\pm256.80~~$&$\chm{\pm138.03~~}$ &  \\
[0.3em]
\multirow{2}{*}{Solar}&$465.29$&$198.94$&$1809.93$&$\bhm{161.80}$&$\chm{161.87}$&$209.95$ & \multirow{2}{*}{$+29.76$}\\
&$\pm216.53~~$&$\pm67.84~~$&$\pm712.45~~$&$\chm{\pm72.72~~}$&$\chm{\pm65.14~~}$&$\pm117.14~~$ &   \\
[0.3em]
\multirow{2}{*}{Birmingham}&$357.26$&$150.64$&$743.63$&$\bhm{111.28}$&$180.13$&$\chm{134.06}$&    \multirow{2}{*}{($\approx 0.00$)} \\ 
&$\pm457.46~~$&$\pm97.01~~$&$\pm480.71~~$&$\chm{\pm37.29~~}$&$\pm101.19~~$&$\chm{\pm99.16~~}$ & \\
[0.3em]
\multirow{2}{*}{PM 2.5}&$1640.79$&$510.57$&$4770.95$&$\bhm{427.16}$&$\chm{427.81}$&$968.07$ & \multirow{2}{*}{$+126.63$} \\
&$\pm1792.22~~$&$\pm220.98~~$&$\pm3784.83~~$&$\chm{\pm259.26~~}$&$\chm{\pm211.78~~}$&$\pm683.55~~$ & \\
[0.3em]
\multirow{2}{*}{Tetouan}&$279.02$&$\bhm{126.62}$&$3715.33$&$168.26$&$187.87$&$\chm{206.24}$ & \multirow{2}{*}{($\approx 0.00$)} \\ 
&$\pm212.56~~$&$\chm{\pm44.07~~}$&$\pm2878.89~~$&$\pm56.81~~$&$\pm46.10~~$&$\chm{\pm350.48~~}$ &  \\
\bottomrule
\end{tabular}
\end{table}

\begin{table}[ht]
\caption{Average cumulative squared error $\widehat{R}(\{f_t\}; \Dte)$ over 30 random trials ($\downarrow$) with datasetting shown in \figref{additional}, comparing offline performances w/o model addition and updates.}
\label{tab:offlinestream}
\centering
\setlength\tabcolsep{14pt}  
\begin{tabular}{c|cccc} \toprule
\multirow{3}{*}{Dataset}&SEGA w/o &Condor w/o &\multirow{2}{*}{FUZZ-CARE w/o}&\multirow{2}{*}{SCSDA} \\ 
&model addition&model addition&\multirow{2}{*}{model updates}&\multirow{2}{*}{(Ours)} \\ 
&for $\Dte$&for $\Dte$&&\\ 
\midrule
\multirow{2}{*}{Seoul bike}&$4405.37$&$1042.30$&$\bhm{595.00}$&$1233.62$ \\
&$\pm6347.40~~$&$\pm474.10~~$&$\chm{\pm209.92~~}$&$\pm1309.01~~$ \\
[0.3em]
\multirow{2}{*}{Bike sharing}&$4691.16$&$1004.57$&$617.09$&$\bhm{219.59}$ \\
&$\pm5591.78~~$&$\pm855.84~~$&$\pm259.34~~$&$\chm{\pm138.03~~}$\\
[0.3em]
\multirow{2}{*}{Solar}&$2020.30$&$448.75$&$\chm{227.00}$&$\bhm{209.95}$ \\
&$\pm2854.70~~$&$\pm174.19~~$&$\chm{\pm107.19~~}$&$\chm{\pm117.14~~}$\\
[0.3em]     
\multirow{2}{*}{Birmingham}&$1698.26$&$441.02$&$213.47$&$\bhm{134.06}$ \\
&$\pm1137.25~~$&$\pm123.03~~$&$\pm122.34~~$&$\chm{\pm99.16~~}$ \\
[0.3em]
\multirow{2}{*}{PM 2.5}&$5607.02$&$891.15$&$\bhm{613.36}$&$968.07$ \\
&$\pm5526.28~~$&$\pm549.98~~$&$\chm{\pm408.51~~}$&$\pm683.55~~$ \\
[0.3em]
\multirow{2}{*}{Tetouan}&$2775.53$&$544.26$&$226.40$&$\bhm{206.24}$ \\
&$\pm2560.64~~$&$\pm229.39~~$&$\pm77.53~~$&$\chm{\pm350.48~~}$\\
\bottomrule
\end{tabular}
\end{table}

\end{document}